\documentclass[sigconf]{acmart}
\AtBeginDocument{%
  }

\copyrightyear{2026}
\acmYear{2026}
\setcopyright{cc}
\setcctype{by-nc-nd}
\acmConference[MM '26]{Proceedings of the 34th ACM International Conference on Multimedia}{November 10--14, 2026}{Rio de Janeiro, Brazil}
\acmBooktitle{Proceedings of the 34th ACM International Conference on Multimedia (MM '26), November 10--14, 2026, Rio de Janeiro, Brazil}
\acmDOI{10.1145/3767308.3836468}
\acmISBN{979-8-4007-2213-4/2026/11}

\usepackage{graphicx}

\usepackage{bbding}
\usepackage[most]{tcolorbox}
\usepackage{booktabs} 
\usepackage{multirow, booktabs, colortbl, xcolor}
\usepackage{pifont}
\usepackage{xcolor}

\usepackage[dvipsnames, svgnames, x11names]{xcolor}
\usepackage{listings}

\usepackage{tabularx}
\usepackage{makecell} 
\usepackage{enumitem}

\begin{document}

\title[Spoken Function Calling]{Spoken Function Calling: A New Perspective on Spoken Language Understanding for Large Audio Language Models}

\author{Yuezhang Peng}
\authornote{Equal Contribution.\quad
\textsuperscript{\dag}Corresponding author.}
\affiliation{%
  \institution{Shanghai Jiao Tong University}
  \city{Shanghai}
  \country{China}
}
\affiliation{%
  \institution{Token Foundry, Alibaba Group}
  \city{Hangzhou}
  \country{China}
}

\author{Yuxin Liu}
\authornotemark[1]
\affiliation{%
  \institution{Shanghai Jiao Tong University}
  \city{Shanghai}
  \country{China}
}

\author{Changfeng Gao}
\affiliation{%
  \institution{Token Foundry, Alibaba Group}
  \city{Beijing}
  \country{China}
}

\author{Zhifu Gao}
\affiliation{%
  \institution{Token Foundry, Alibaba Group}
  \city{Hangzhou}
  \country{China}
}

\author{Qian Chen}
\affiliation{%
  \institution{Token Foundry, Alibaba Group}
  \city{Hangzhou}
  \country{China}
}

\author{Xiangang Li}
\affiliation{%
  \institution{Token Foundry, Alibaba Group}
  \city{Hangzhou}
  \country{China}
}

\author{Xie Chen}
\authornotemark[2]
\affiliation{%
  \institution{Shanghai Jiao Tong University}
  \city{Shanghai}
  \country{China}
}
\affiliation{%
  \institution{Shanghai Innovation Institute}
  \city{Shanghai}
  \country{China}
}

\renewcommand{\shortauthors}{Yuezhang Peng et al.}

\begin{abstract}
Spoken Language Understanding (SLU) is the core component of task-oriented dialogue systems and a pivotal link in achieving seamless human-agent interaction. While traditional SLU can effectively extract user semantics for closed-set tasks after in-domain supervised fine-tuning, it faces significant challenges in leveraging in-context learning for open-domain tasks due to its ambiguous rule definitions. This work proposes Spoken Function Calling (SFC), a novel semantic understanding perspective that optimizes semantic understanding with structured rule definitions, to evolve beyond traditional closed-set SLU. Specifically, we curate and extend a suite of spoken functions based on traditional SLU datasets, construct a multi-agent system to synthesize the SFC-Bench dataset, evaluate the performance of Large Language Models (LLMs) and Large Audio Language Models (LALMs), and enhance the SFC capabilities of LALMs through post-training. Experiments demonstrate that SFC outperforms traditional SLU, substantially enhancing the semantic extraction accuracy for LLMs and LALMs.\footnote{Code and data: \url{https://github.com/QwenAudio/FunResearch/tree/main/SpokenFC}.}
\end{abstract}




\maketitle

\section{Introduction}

\begin{figure}[!t]
    \setlength{\abovecaptionskip}{3pt}
    \centering 
    \includegraphics[width=0.75\linewidth]{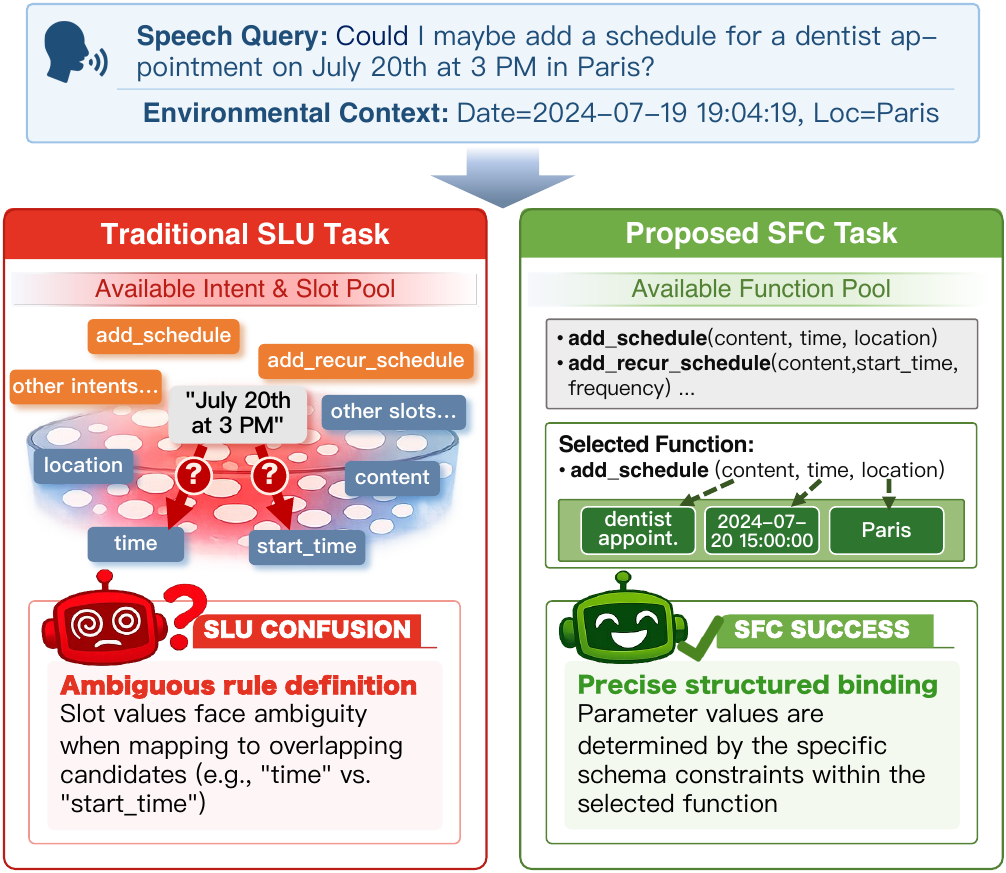} 
    \caption{Comparison of traditional SLU and proposed SFC. SFC resolves parameter ambiguity through a precise and structured function definition.}
    \label{fig:comparison}
    \vspace{-10pt}
\end{figure}

\begin{figure*}[!t]
    \centering 
    \includegraphics[width=0.9\textwidth]{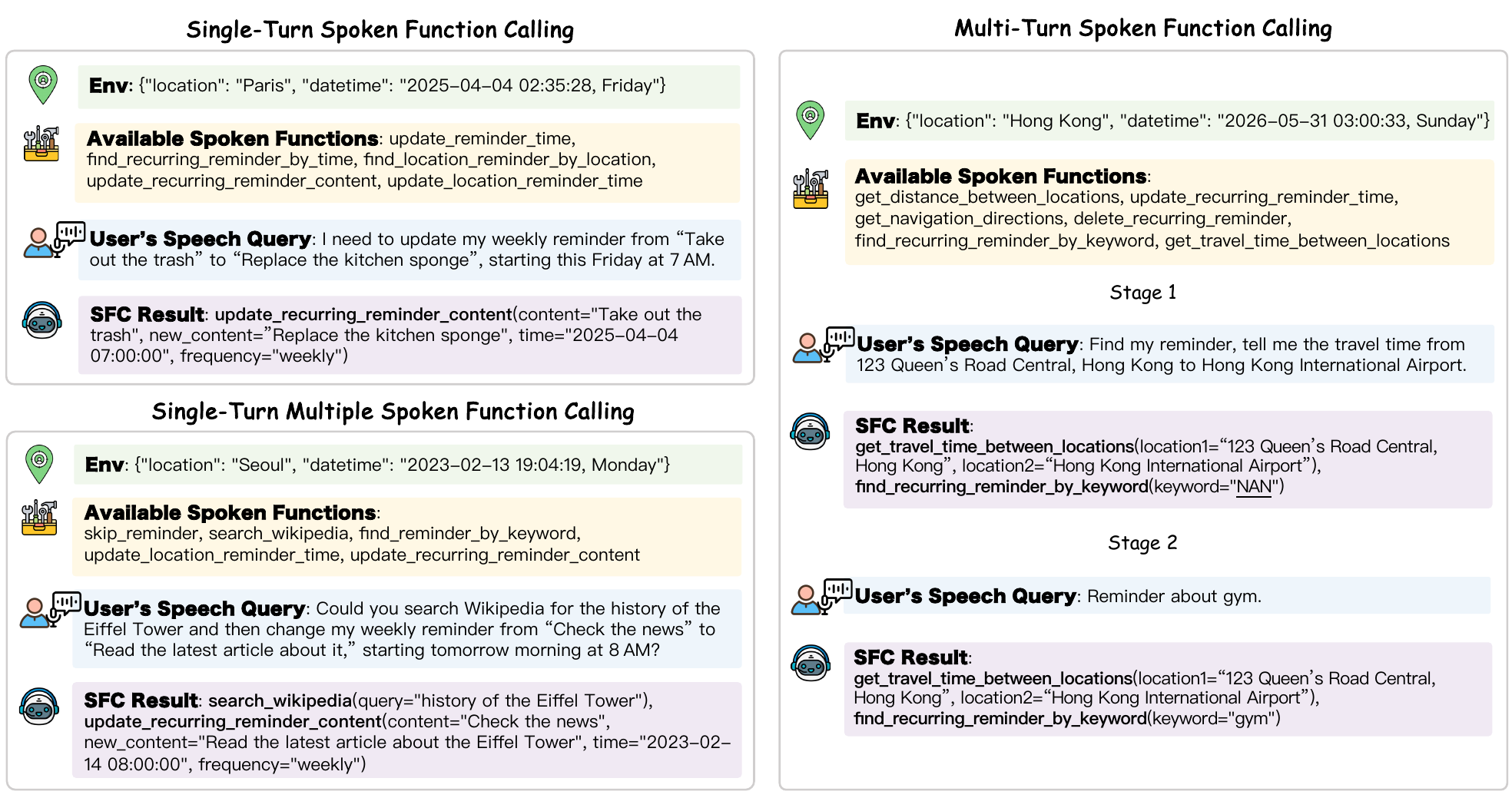} 
    \caption{Example of Multi-Intent and Multi-Turn Spoken Function Calling tasks.}
    \label{fig:sfc_example}
    \vspace{-10pt}
\end{figure*}

Spoken Language Understanding (SLU) parses the semantic information of user queries through intent classification and slot filling to facilitate the execution of downstream tasks \cite{tur2011spoken}. This paradigm has been extensively deployed in voice assistants across smartphones, smart homes, and automotive systems \cite{lugosch2019speech,bastianelli2020slurp,peng2025mac}. However, traditional SLU tasks operate under the assumption that models possess prior knowledge of all predefined intent and slot schemas. While models can memorize these rules via in-domain Supervised Fine-Tuning (SFT) to achieve superior performance, they struggle with open vocabulary and out-of-domain tasks \cite{peng2025mac}. Even state-of-the-art (SOTA) Large Language Models (LLMs) and Large Audio-Language Models (LALMs) fail to achieve satisfactory results on such tasks through In-Context Learning (ICL) \cite{brown2020language}. In this context, we define ``open-domain" as the capability of a model to understand and execute tasks through ICL based on dynamic functional descriptions, rather than relying on in-domain SFT to pre-memorize SLU schemas.

Recently, function calling has emerged as a transformative paradigm for semantic parsing and task execution \cite{patil2024gorilla, paranjape2023art}. Unlike traditional SLU, which requires agents to pre-memorize rigid rules, function calling enables LLMs to interpret function definitions dynamically through ICL and structured prompts. This approach facilitates the autonomous invocation of the Model Context Protocol (MCP) \cite{hou2025model} for diverse open-domain tasks, bypassing the need for SFT on specific functions. Inspired by the success of function calling in open-domain scenarios, we investigate a pivotal question: \textbf{Can Spoken Function Calling (SFC) evolve beyond traditional SLU to enhance the semantic extraction performance of agents?}

However, addressing this question presents a critical challenge: \textbf{the lack of SFC datasets that correspond to traditional SLU tasks}. Specifically, existing LLMs function call datasets focus on complex reasoning within code or mathematics, which does not align with the requirements of traditional SLU. Furthermore, queries and tools designed for text-centric tasks lack the inherent uncertainty and linguistic flexibility of human speech.

In this work, we leverage SLU datasets and a multi-agent system to construct SFC-Bench, the first large-scale SFC dataset. Through comparative experiments, we demonstrate that SFC is a more suitable perspective for LALM agents than traditional SLU, as shown in Figure \ref{fig:comparison}. We further establish a benchmark to investigate the performance and limitations of LLMs and LALMs on SFC tasks of varying levels. In summary, our key contributions are listed below.


\begin{itemize}[leftmargin=*]
    \item \textbf{Perspective Shift}: We propose \textbf{Spoken Function Calling} as a novel perspective for task-oriented spoken semantics extraction, to extend traditional closed-set SLU based on intent classification and slot filling in the agent era.
    \item \textbf{Post-training Optimization}: We investigate reinforcement learning (RL) based post-training strategies for LALMs and propose a \textbf{fine-grained reward} method. Leveraging these strategies, we develop \textbf{SpokenFC-7B} through post-training on a comprehensive synthetic dataset.
    \item \textbf{Dataset Construction}: We introduce \textbf{SFC-Bench}, the first large-scale SFC dataset. Starting from a curated tool set of 300 spoken functions derived from existing SLU datasets, we employ a multi-agent system to synthesize multi-level queries and labels.
    \item \textbf{Empirical Validation}: Our experiments confirm that: 1) SFC serves as a more effective semantic understanding method for LALMs compared to traditional SLU. 2) Even SOTA LALMs still face challenges in high-level SFC tasks involving multi-intent and ambiguous semantics. 3) SpokenFC-7B outperforms powerful closed-source models on SFC tasks while maintaining robust performance on speech recognition and audio reasoning.

\end{itemize}


\section{Related Work}

\subsection{Traditional SLU}

\subsubsection{LLMs for SLU} 
LLMs pre-trained under the scaling law \cite{kaplan2020scaling} exhibit emergent reasoning and generalization capabilities across various Natural Language Processing (NLP) tasks, successfully addressing complex problems in question-answering \cite{zhuang2023toolqa}, mathematics \cite{lu2023mathvista}, and coding \cite{chen2021evaluating}. Recent research explores the application of LLMs to SLU tasks. Early research \cite{he2023can} investigates using ICL for intent classification and slot filling, demonstrating that while LLMs perform commendably in intent classification, they still face challenges with the more complex structural definitions of slot filling. Whisma \cite{li2024whisma} employs a Whisper \cite{radford2023robust} encoder and Llama-3 \cite{dubey2024llama} decoder with modality alignment training, enabling zero-shot SLU capabilities that outperform traditional pipeline methods. Furthermore, MAC-SLU \cite{peng2025mac} establishes a unified benchmark for LLMs and LALMs on SLU, indicating that while current LLMs and LALMs can perform well on sub-tasks via ICL (e.g., isolated intent classification or slot filling), their overall accuracy remains low and significantly lags behind models optimized through SFT.

\subsubsection{SLU Datasets} SLU datasets primarily include single-intent benchmarks such as ATIS, SNIPS, FSC, and SLURP \cite{hemphill1990atis, coucke2018snips, lugosch2019speech,bastianelli2020slurp}, as well as multi-intent datasets like MixATIS and MixSNIPS \cite{qin2020agif}. The ATIS \cite{hemphill1990atis} dataset focuses on simple voice queries for flight information with only 16 intent categories. SNIPS \cite{coucke2018snips} and FSC \cite{lugosch2019speech} target smart home scenarios but remain limited in scope, featuring only 7 and 13 intents, respectively. The simplicity and homogeneity of ATIS, SNIPS, and FSC allow even non-LLM-based SLU models to achieve over 95\% overall accuracy \cite{qin2021co}. SLURP \cite{bastianelli2020slurp} is currently the most widely used large-scale SLU dataset, expanding the intent categories to 46. In the context of Chinese SLU, MAC-SLU \cite{peng2025mac} is derived from text commands in the automotive cabin with speech synthesized via Text-to-Speech (TTS) models. MAC-SLU comprises 8 domains, 81 intents, and 192 slots, including multi-intent data with up to five concurrent intents, surpassing previous datasets in both diversity and complexity.
Furthermore, although dialogue state tracking \cite{Williams2013TheDS, Williams2016TheDS} is inherently linked with SLU, and existing benchmarks like MTOP \cite{li2021mtop} and MultiWOZ \cite{budzianowski2018multiwoz} have adopted function-calling structures, their scope is restricted to text-based, multi-turn dialogues. In contrast, this work concentrates on end-to-end semantic understanding specifically for the speech modality and the precise invocation of spoken tools, presenting fundamental differences in both input modality and task objectives.


\subsection{Function Calling}

\subsubsection{LLMs for Function Calling} Function calling empowers LLMs with tool-use capabilities to overcome inherent limitations such as hallucinations \cite{ji2023towards} and the lack of real-time information. ReAct \cite{yao2022react} represents a milestone in function calling, proposing an interleaved reasoning and acting paradigm. This approach guides models to generate reasoning traces before executing an action, leveraging Chain-of-Thought (CoT) \cite{wei2022chain} to plan tool calls and enhancing the model's ability to solve complex problems. Toolformer \cite{schick2023toolformer} introduces a self-supervised learning method that automatically inserts API calls into text and filters low-quality samples, teaching the model how to invoke external APIs without extensive human annotation. ToolRL \cite{qian2025toolrl} utilizes RL training to bolster the reasoning capabilities of LLMs during function calling, achieving performance superior to both base and SFT models.

\subsubsection{Function Calling Datasets} Current function calling datasets include the Berkeley Function Calling Leaderboard (BFCL) \cite{patil2024gorilla}, API-Bank \cite{li2023api}, ToolBench \cite{qin2023toolllm}, and ToolACE \cite{liu2024toolace}. BFCL \cite{patil2024gorilla} evaluates function calling using a comprehensive dataset encompassing HuggingFace, TorchHub, and TensorHub APIs. API-Bank \cite{li2023api} provides the first benchmark specifically for tool-augmented LLMs, covering capabilities ranging from single-turn queries to multi-turn dialogue management. ToolLLM \cite{qin2023toolllm} constructs ToolBench, a large-scale instruction-tuning dataset featuring over 16,000 functional RESTful APIs. ToolACE \cite{liu2024toolace} utilizes an automated agentic pipeline to generate accurate, complex, and diverse tool-learning data, enhancing function calling through data augmentation. Despite the proliferation of these datasets, they remain focused on code, mathematics, and complex reasoning, which do not align with spoken semantic understanding scenarios. Furthermore, speech-centric datasets like StepEval-Audio-Toolcall \cite{wu2025step} support only four categories of tools, which is insufficient for a comprehensive evaluation of LALMs' performance in SFC tasks.

\begin{figure*}[!t]
    \centering 
    \includegraphics[width=0.9\textwidth]{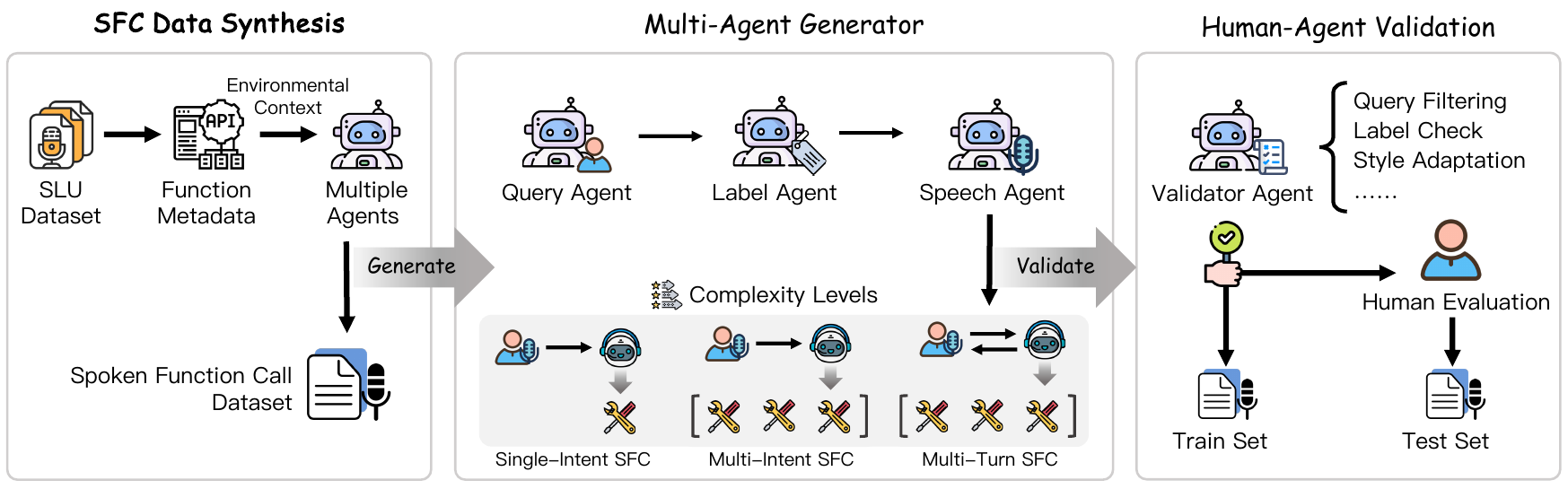} 
    \caption{Multi-Agent System for Data Construction.}
    \label{fig:multi-agent-system}
    \vspace{-10pt}
\end{figure*}

\section{Methodology}

\subsection{Spoken Function Calling}

Within the paradigm of ICL for LLMs and LALMs, the SLU based on intent classification and slot filling can be conceptualized as a semantic extraction and generation process grounded in instruction following. Its task mapping relationship can be defined as:
\begin{equation}
F_{\text{SLU}} : (A, S_{\text{int}}, S_{\text{slot}}) \rightarrow Y_{\text{SLU}},
\end{equation}
where $A$ denotes the input raw speech signal, $S_{\text{int}}$ and $S_{\text{slot}}$ denote the predefined sets of candidate intent and slot labels, respectively. The output $Y_{\text{SLU}}$ is represented as a set of semantic frames:
\begin{equation}
Y_{\text{SLU}} = \{I_1, \dots, I_n, (s_1, v_1), \dots, (s_m, v_m)\}.
\end{equation}
In this structure, $I_i \in S_{\text{int}}$ is the $i$-th identified intent label, while $s_j \in S_{\text{slot}}$ and $v_j$ represent the corresponding slot name and its associated value, respectively.

To accommodate the requirements for agents to handle open-domain SLU with customized intents and slots and to circumvent the reliance on memorizing rigid rules through SFT, this work proposes a novel perspective: Spoken Function Calling. SFC aims to map spoken instructions into a sequence of directly executable structured API calls:
\begin{equation}
F_{\text{SFC}} : (A, E, D_{\text{func}}) \rightarrow Y_{\text{SFC}}.
\end{equation}
In this definition, $E$ explicitly introduces the environmental context, which serves to compensate for semantic omissions common in spoken language; $D_{\text{func}}$ represents the function definition schema following the OpenFunction protocol. The detailed output $Y_{\text{SFC}}$ is manifested as a set of structured function calls:
\begin{equation}
Y_{\text{SFC}} = \{ f_i(k_{i,1} = v_{i,1}, k_{i,2} = v_{i,2}, \dots, k_{i,m_i} = v_{i,m_i}) \}_{i=1}^n,
\end{equation}
where $f_i$ is the retrieved function name, and $k_{i,j}$ and $v_{i,j}$ represent parameter keys and values. If a parameter is missing, it is assigned a value of \texttt{NAN} to trigger subsequent completion mechanisms. Compared to traditional SLU, the SFC perspective offers the following significant advantages: \textbf{1) Robustness of Structured Constraints}: The schema constraints introduced via $D_{\text{func}}$ serve as strong prior knowledge, effectively resolving the boundary ambiguity issues encountered by traditional SLU when performing complex parameter extraction. \textbf{2) Coupled Intent and Slot Parsing}: The function definition $D_{\text{func}}$ simultaneously constrains the parsing of both function names (intents) and parameter values (slots). This addresses a critical limitation in traditional SLU, where the correlation between intents and slots is difficult to define explicitly.

\subsection{RL-based Post-training}

This study adopts Group Relative Policy Optimization (GRPO) \cite{shao2024deepseekmath} as the core algorithm for post-training. Compared to the traditional Proximal Policy Optimization (PPO) \cite{schulman2017proximal} algorithm, a significant advantage of GRPO is that it eliminates the need to train an additional large-scale value function model and substantially reduces the memory footprint and computational overhead.

For each spoken query $q$, GRPO samples a group of outputs $\{o_1, o_2, \dots, o_G\}$ from the old policy $\pi_{\theta_{\text{old}}}$. The original objective function $J_{\text{GRPO}}(\theta)$ is defined as follows:
\begin{equation}
\begin{split}
J_{\text{GRPO}}(\theta) = \mathbb{E}_{q \sim P(Q), \{o_i\}_{i=1}^G \sim \pi_{\theta_{\text{old}}}} \bigg[ \frac{1}{G} \sum_{i=1}^G \Bigl( \min \Bigl( \tfrac{\pi_\theta(o_i|q)}{\pi_{\theta_{\text{old}}}(o_i|q)} \hat{A}_i, \\
\text{clip} \left( \tfrac{\pi_\theta(o_i|q)}{\pi_{\theta_{\text{old}}}(o_i|q)}, 1-\epsilon, 1+\epsilon \right) \hat{A}_i \Bigr) - \beta D_{\text{KL}}(\pi_\theta \| \pi_{\text{ref}}) \Bigr) \bigg].
\end{split}
\end{equation}
The advantage function $\hat{A}_{i}$ for each output $o_i$ is computed by normalizing the rewards within each group:
\begin{equation}
\hat{A}_{i} = \frac{R^i - \text{mean}(\mathbf{R})}{\text{std}(\mathbf{R})},
\end{equation}
where $\mathbf{R} = \{R^1, R^2, \dots, R^G\}$ denotes the set of rewards for the current sampling group. RL for LLMs typically employs an exact match (EM) reward. To precisely characterize the structured features of SFC, we represent the generated sequence $Y_{\text{gen}}$ and the ground-truth sequence $Y_{\text{gt}}$ as sets of function calls:
\begin{equation}
R_{\text{EM}} = \mathbb{I}(Y_{\text{gen}} = Y_{\text{gt}}),
\end{equation}
where model output and ground truth are respectively defined as:
\begin{equation}
\begin{cases}
Y_{\text{gen}} = \{ \hat{f}_i (\hat{k}_{i,1} = \hat{v}_{i,1}, \hat{k}_{i,2} = \hat{v}_{i,2}, \dots) \}_{i=1}^{\hat{n}} \\
Y_{\text{gt}} = \{ f_j (k_{j,1} = v_{j,1}, k_{j,2} = v_{j,2}, \dots) \}_{j=1}^n.
\end{cases}
\end{equation}
In this context, $\mathbb{I}(\cdot)$ is the indicator function, while $\hat{f}, \hat{k}, \text{ and } \hat{v}$ denote the generated function name, parameter key, and parameter value, respectively. Although the EM reward helps prevent reward hacking, it is overly restrictive for SFC tasks involving complex JSON structures, as it requires total identity across all components. This often leads to sparse rewards during early training stages, as the model struggles to hit all structured elements simultaneously, thereby causing slow convergence.

\begin{figure*}[!t]
    \centering

    \begin{minipage}[c]{0.44\textwidth}
        \centering
        \includegraphics[width=\linewidth]
            {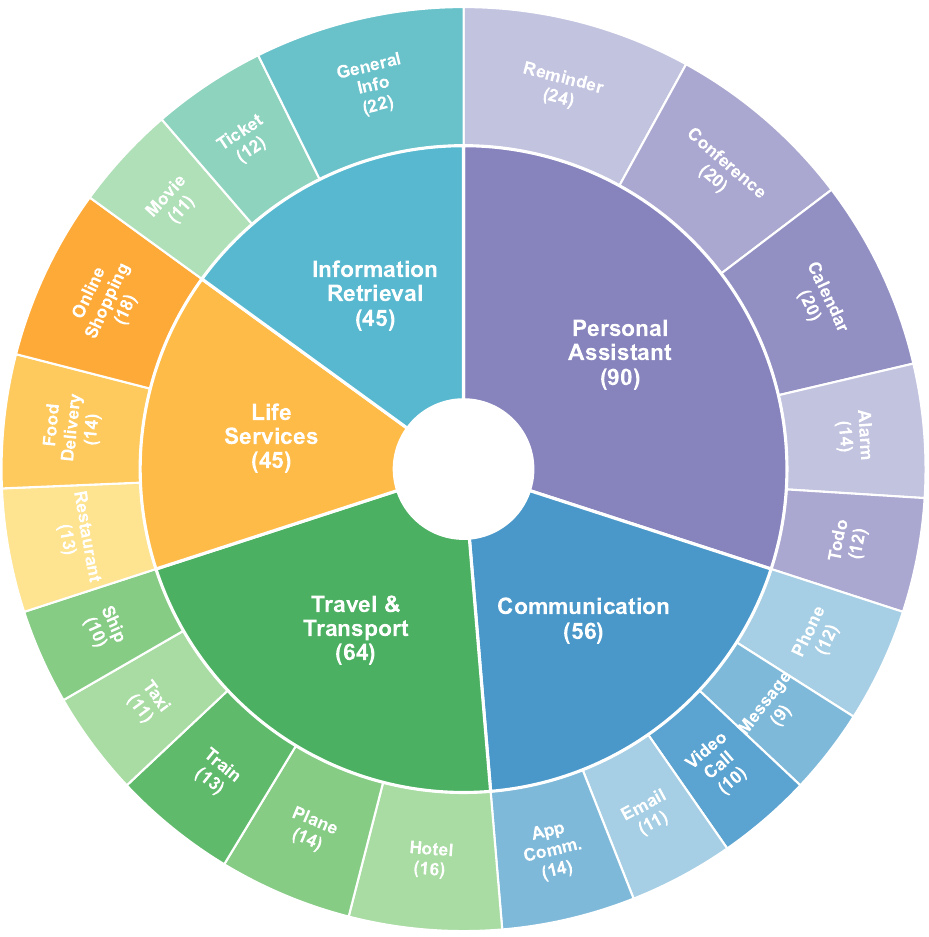}
    \end{minipage}
    \hspace{0.08\textwidth} 
    \begin{minipage}[c]{0.344\textwidth}
        \centering
        \renewcommand{\arraystretch}{1.2}
        \setlength{\tabcolsep}{6pt}

        \resizebox{\linewidth}{!}{
            \begin{tabular}{lccc}
                \toprule
                \textbf{Metrics}
                & \textbf{Train}
                & \textbf{Test-ID}
                & \textbf{Test-OOD} \\
                \midrule

                \multicolumn{4}{l}{
                    \cellcolor{gray!15}\textbf{Level 1}
                } \\
                Avg. Question Length & 774.58 & 728.41 & 704.69 \\
                Avg. Answer Length   & 20.68  & 18.35  & 13.34  \\
                Avg. Audio Length    & 7.30   & 6.62   & 6.69   \\
                \midrule

                \multicolumn{4}{l}{
                    \cellcolor{gray!15}\textbf{Level 2}
                } \\
                Avg. Question Length & 708.80 & 707.85 & 744.68 \\
                Avg. Answer Length   & 30.14  & 28.51  & 26.93  \\
                Avg. Audio Length    & 10.65  & 10.28  & 10.21  \\
                \midrule

                \multicolumn{4}{l}{
                    \cellcolor{gray!15}\textbf{Level 3-1}
                } \\
                Avg. Question Length & 792.51 & 808.55 & 732.21 \\
                Avg. Answer Length   & 40.92  & 44.93  & 27.68  \\
                Avg. Audio Length    & 8.70   & 8.65   & 7.66   \\
                \midrule

                \multicolumn{4}{l}{
                    \cellcolor{gray!15}\textbf{Level 3-2}
                } \\
                Avg. Question Length & 827.57 & 806.98 & 793.87 \\
                Avg. Answer Length   & 73.96  & 63.17  & 52.71  \\
                Avg. Audio Length    & 13.15  & 11.55  & 11.32  \\
                \bottomrule
            \end{tabular}
        }
    \end{minipage}

    \caption{The domain distribution (left) and detailed statistics
    (right) of average question lengths (tokens), answer lengths
    (tokens), and audio lengths (seconds) across different levels
    of splits.}
    \label{fig:combined_data_stats}
\end{figure*}

To mitigate this, we propose a \textbf{Fine-Grained reward} method by decomposing the SFC task into three core sub-tasks: function name recognition, parameter key matching, and parameter value extraction. By providing positive feedback when a sub-task is performed correctly, we incentivize the model to master individual components before achieving full function call synthesis. The final reward $R_{\text{Fine-Grained}}$ is normalized to the $[-1, 1]$ range by scaling the arithmetic mean of the three dimensions:
\begin{equation}
R_{\text{Fine-Grained}} = 2 \cdot \left( \frac{r_{\text{name}} + r_{\text{keys}} + r_{\text{values}}}{3} \right) - 1
\end{equation}
\textbf{Function Name Reward ($r_{\text{name}}$)}: Evaluates the accuracy of intent recognition. It is measured by calculating the match ratio between the generated function names $\hat{f}_i$ and the ground truth $f_i$:
    \begin{equation}
    r_{\text{name}} = \frac{1}{n} \sum_{i=1}^n \mathbb{I}(\hat{f}_i = f_i).
    \end{equation} 
\textbf{Parameter Key Matching Reward ($r_{\text{keys}}$)}: Measures the model's command over the tool parameter structure. Let $\hat{K} = \{\hat{k}_j\}$ be the set of generated keys and $K = \{k_j\}$ be the ground-truth keys. This is calculated using the Jaccard similarity:
    \begin{equation}
    r_{\text{keys}} = \frac{|\hat{K} \cap K|}{|\hat{K} \cup K|}.
    \end{equation}
\textbf{Parameter Value Accuracy Reward ($r_{\text{values}}$)}: Evaluates the semantic consistency of parameter values for keys present in the intersection of $\hat{K}$ and $K$. To provide denser feedback, we employ a fine-grained metric based on the normalized edit distance:
    \begin{equation}
    r_{\text{values}} = \frac{\sum_{k \in \hat{K} \cap K} (1 - \textit{EditDistance}(\hat{v}_k, v_k))}{|K|}.
    \end{equation}

\section{Spoken Function Calling Dataset}

\subsection{Spoken Functions Collection}


The construction of the spoken function meta-dataset followed a structured three-step pipeline. \textbf{Firstly}, we aggregated a comprehensive set of scenarios from widely utilized SLU benchmarks. \textbf{Secondly}, we leveraged Gemini-2.5 Pro to transform these scenarios into standardized function definitions. To ensure functional diversity and logical complexity, we referenced established text-based function-calling benchmarks such as BFCL \cite{patil2024gorilla} and ToolACE \cite{liu2024toolace}. The generated functions cover a broad spectrum of operations, ranging from standard CRUD (Create, Read, Update, Delete) actions to domain-specific complex logic. \textbf{Thirdly}, we conducted manual Refinement, where every generated function underwent rigorous human filtering and correction.


In total, we curated 300 daily-life oriented spoken tools spanning 5 domains and 21 scenarios, as presented in Figure \ref{fig:combined_data_stats} (left). To evaluate the model's generalization capabilities, we partitioned the function set into 232 In-Distribution (ID) and 68 Out-of-Distribution (OOD) functions. Although the diversity and inherent complexity of the Test-OOD set are slightly lower than those of the Test-ID set due to this distribution, it does not compromise the validity of the experiments in assessing the model's training generalization. All subsequent training was conducted solely on data synthesized from the ID subset, while evaluations were performed on both ID and OOD test sets to verify robustness and transferability.



\subsection{SFC-Bench Construction}

\newcommand{\cmark}{\textcolor{green!70!black}{\ding{51}}} 
\newcommand{\xmark}{\textcolor{red}{\ding{55}}} 


\begin{table}[!t]
    \small
    \centering
    \renewcommand{\arraystretch}{0.95}
    \setlength{\tabcolsep}{4pt}
    \setlength{\aboverulesep}{1.5pt}
    \setlength{\belowrulesep}{1.5pt}
    \caption{Sample distribution in the SFC-Bench dataset.}
    \label{sfc-tab:dataset_distribution}
    \resizebox{\columnwidth}{!}{%
    \begin{tabular}{l c c c c c}
        \toprule
        \textbf{Level} & \textbf{Train} & \textbf{Test-ID}
        & \textbf{Test-OOD}
        & \makecell[c]{\textbf{Multi-}\\\textbf{Intent}}
        & \makecell[c]{\textbf{Multi-}\\\textbf{Turn}} \\
        \midrule
        L1    & 2,647 & 902   & 321 & \xmark & \xmark \\
        L2    & 817   & 480   & 94  & \cmark & \xmark \\
        L3-1  & 447   & 317   & 126 & \xmark & \cmark \\
        L3-2  & 588   & 287   & 52  & \cmark & \cmark \\
        \midrule
        Total & 4,499 & 1,986 & 593 & -- & -- \\
        \bottomrule
    \end{tabular}%
    }
    \vspace{-10pt}
\end{table}

\textbf{Workflow Overview.} The workflow for SFC-Bench construction is depicted in Figure \ref{fig:multi-agent-system}. We synthesize the dataset by leveraging function metadata curated from SLU benchmarks, integrated with diverse environmental contexts to facilitate data augmentation. 

\textbf{Multi-Agent Generator.} Central to this workflow is the Multi-Agent Generator, powered by GPT-OSS-120B \cite{agarwal2025gpt} with custom system prompts, which comprises three primary functional agents: \textbf{1) Query Agent}: Generates text queries based on pre-selected functions and environmental context. For Level 3 data involving ambiguous semantics, an additional step of fuzzy rewriting is performed. \textbf{2) Label Agent}: Produces accurate function calls based on the chosen functions, environmental context, and queries. For ambiguous Level 3 queries, it generates function calls containing \texttt{NAN} parameters to assess potential hallucination. \textbf{3) Speech Agent}: Introduces spoken language features (e.g., redundancy, self-corrections) and verbalizes technical formats (e.g., converting ``\textit{abc@gmail.com}" to ``\textit{abc at gmail dot com}"). Finally, it employs IndexTTS-2 \cite{zhou2025indextts2} to synthesize speech queries by utilizing speaker timbre profiles from the LibriSpeech dataset \cite{panayotov2015librispeech} to mimic diverse vocal characteristics. To ensure experimental rigor, distinct speaker identities are assigned to the training, validation, and test sets, maintaining strict speaker isolation across the data splits.

\begin{table*}[!t]\small
    \centering
    \renewcommand{\arraystretch}{1.1}
    \footnotesize
    \caption{Qualitative comparison between SFC and traditional SLU tasks. The Result column indicates the success (\cmark) or failure (\xmark) of each task based on the ground truth. \textcolor{RoyalBlue}{Blue} text highlights specific discrepancies.}
    \resizebox{\textwidth}{!}{
    \begin{tabular}{l p{6.5cm} p{6.5cm} c}
        \toprule
        \textbf{Task} & \textbf{Model Prediction} & \textbf{Ground Truth} & \textbf{Result} \\
        \midrule
        
        \multicolumn{4}{l}{\cellcolor{gray!15}\textbf{Case 1: Hey, delete my doctor appointment reminder for tomorrow morning at 8 AM, I'm in a rush!}} \\
        \midrule
        \textbf{SFC} & \texttt{delete\_reminder}(content = "doctor appointment", time = "2025-03-20 08:00:00") & \texttt{delete\_reminder}(content = "doctor appointment", time = "2025-03-20 08:00:00") & \cmark \\
        \textbf{SLU} & "intent": "delete\_reminder", "slots": \{ "content": "doctor appointment" \} & "intent": "delete\_reminder", "slots": \{ "content": "doctor appointment", \textcolor{RoyalBlue}{"time": "2025-03-20 08:00:00"} \} & \xmark \\
        \midrule

        \multicolumn{4}{l}{\cellcolor{gray!15}\textbf{Case 2: Um, could I maybe add a schedule for a dentist appointment on July 20th at 3 PM in Paris?}} \\
        \midrule
        \textbf{SFC}& \texttt{add\_schedule}(content = "dentist appointment", time = "2024-07-20 15:00:00", location="Paris") & \texttt{add\_schedule}(content = "dentist appointment", time = "2024-07-20 15:00:00", location="Paris") & \cmark \\
        \textbf{SLU} & "intent": "add\_schedule", "slots": \{ "content": "dentist appointment", \textcolor{RoyalBlue}{"start\_time"}: "2024-07-20 15:00:00", "location": "Paris" \} & "intent": "add\_schedule", "slots": \{ "content": "dentist appointment", \textcolor{RoyalBlue}{"time"}: "2024-07-20 15:00:00", "location": "Paris" \} & \xmark \\
        \bottomrule
    \end{tabular}}
    \label{tab:qualitative_comparison}
\end{table*}

\begin{table}[!t]
\footnotesize
\centering
\renewcommand{\arraystretch}{1}
\setlength{\tabcolsep}{3pt}
\setlength{\aboverulesep}{1pt}
\setlength{\belowrulesep}{1pt}
\setlength{\cmidrulesep}{1pt}

\caption{Experimental results of SLU and SFC tasks. The metrics include Intent (Function Name) accuracy, Slot (Parameter Value) F1 Score, and Overall Accuracy across ID and OOD test sets. Best results are in \textbf{bold}.}
\label{sfc-tab:combined_results}

\resizebox{\columnwidth}{!}{%
\begin{tabular}{ll ccc ccc}
\toprule
& & \multicolumn{3}{c}{\textbf{Test-ID}}
  & \multicolumn{3}{c}{\textbf{Test-OOD}} \\
\cmidrule(lr){3-5}\cmidrule(lr){6-8}
& \textbf{Task}
& \textbf{Intent} & \textbf{Slot} & \textbf{Overall}
& \textbf{Intent} & \textbf{Slot} & \textbf{Overall} \\
\midrule

\multicolumn{8}{l}{\cellcolor{gray!15}\textbf{Qwen3-8B, Text Input}} \\
\multirow{2}{*}{ICL}
& SLU & 96.01 & 85.20 & 69.73 & 97.20 & 92.52 & 83.80 \\
& SFC & \textbf{97.56} & \textbf{91.83} & \textbf{85.70}
      & \textbf{99.07} & \textbf{95.83} & \textbf{91.90} \\
\cmidrule(lr){1-8}
\multirow{2}{*}{SFT}
& SLU & 96.74 & 84.26 & 70.44 & 96.88 & 90.24 & 82.24 \\
& SFC & \textbf{97.67} & \textbf{92.68} & \textbf{87.03}
      & \textbf{99.07} & \textbf{94.48} & \textbf{90.03} \\
\midrule

\multicolumn{8}{l}{\cellcolor{gray!15}\textbf{Qwen2.5-7B, Text Input}} \\
\multirow{2}{*}{ICL}
& SLU & \textbf{93.90} & 78.73 & 61.86 & 93.15 & 82.71 & 70.57 \\
& SFC & 91.57 & \textbf{85.41} & \textbf{73.95}
      & \textbf{93.46} & \textbf{87.58} & \textbf{75.70} \\
\cmidrule(lr){1-8}
\multirow{2}{*}{SFT}
& SLU & 95.79 & 84.94 & 68.63 & 93.77 & 87.42 & 77.00 \\
& SFC & \textbf{96.78} & \textbf{90.97} & \textbf{83.15}
      & \textbf{99.38} & \textbf{94.48} & \textbf{89.72} \\
\midrule

\multicolumn{8}{l}{\cellcolor{gray!15}\textbf{Qwen2.5-Omni-7B, Speech Input}} \\
\multirow{2}{*}{ICL}
& SLU & \textbf{94.46} & 68.06 & 53.88 & 95.64 & 69.12 & 50.16 \\
& SFC & 93.02 & \textbf{79.69} & \textbf{63.86}
      & \textbf{97.20} & \textbf{80.43} & \textbf{68.54} \\
\cmidrule(lr){1-8}
\multirow{2}{*}{SFT}
& SLU & 98.56 & 85.50 & 75.61 & 96.57 & 74.79 & 57.01 \\
& SFC & \textbf{99.45} & \textbf{87.80} & \textbf{80.71}
      & \textbf{100.0} & \textbf{79.52} & \textbf{67.29} \\
\bottomrule
\end{tabular}%
}
\vspace{-5pt}
\end{table}

\begin{table}[!t]\small
    \centering
    \caption{Comparison of average confidence (\textit{Log Probabilities}) between SFC and SLU tasks.}
    \begin{tabular}{l c c}
        \toprule
         & \textbf{Test-ID} & \textbf{Test-OOD} \\
        \midrule
        SFC Confidence & \textbf{-1.62} & \textbf{-1.57} \\
        SLU Confidence & -1.87 & -1.72 \\
        
        \bottomrule
    \end{tabular}
    \label{fig:log_prob}
    \vspace{-10pt}
\end{table}

\textbf{Validator.} Data synthesized by the Multi-Agent Generator undergoes inspection and filtering by the validator agent to ensure logical and grammatical soundness, as well as consistency between queries and labels. For the test set, an additional human validation phase is conducted to guarantee quality. The distribution of the SFC-Bench is summarized in Table \ref{sfc-tab:dataset_distribution}. We generated over 7,000 SFC data samples, covering a range of Multi-Intent and Multi-Turn scenarios.

Statistical data in Figure \ref{fig:combined_data_stats} (right) demonstrate a trend in complexity as difficulty levels escalate. The average answer length surges from 20 tokens at Level 1 to over 70 tokens at Level 3-2, highlighting the stringent requirements for structured parsing in multi-intent and multi-turn semantic completion tasks. Concurrently, the mean audio duration increases from about 7 seconds to 13 seconds. This increased audio load substantially exacerbates the challenges of ASR error propagation and long-range semantic alignment, which constitute the primary obstacles to maintaining robustness in LALMs when processing complex instructions.


\begin{table*}[!t]
    \centering
    \setlength{\cmidrulewidth}{0.05mm}
    \renewcommand{\arraystretch}{1.1}
    \setlength{\tabcolsep}{3pt} 
    \caption{Experimental results on the SFC Test-ID. The metrics include Intent (Function Name) Accuracy, Slot (Parameter Value) F1 Score, and Overall Accuracy. Best results are in \textbf{bold}.}
    \label{tab:test_id_en}
    \resizebox{\textwidth}{!}{
    \begin{tabular}{ll p{2mm} ccc p{2mm} ccc p{2mm} ccc p{2mm} ccc p{2mm} ccc}
        \toprule
        & & & \multicolumn{3}{c}{\textbf{Level 1}} & & \multicolumn{3}{c}{\textbf{Level 2}} & & \multicolumn{3}{c}{\textbf{Level 3-1}} & & \multicolumn{3}{c}{\textbf{Level 3-2}} & & \multicolumn{3}{c}{\textbf{Average (Turn-Level)}} \\
        \cmidrule(lr){4-6} \cmidrule(lr){8-10} \cmidrule(lr){12-14} \cmidrule(lr){16-18} \cmidrule(lr){20-22}
        
        \textbf{Category} & \textbf{Model} && \textbf{Intent} & \textbf{Slot} & \textbf{Overall} && \textbf{Intent} & \textbf{Slot} & \textbf{Overall} && \textbf{Intent} & \textbf{Slot} & \textbf{Overall} && \textbf{Intent} & \textbf{Slot} & \textbf{Overall} && \textbf{Intent} & \textbf{Slot} & \textbf{Overall} \\
        \midrule
        
        \multirow{4}{*}{\textbf{LLM}} 
        & Llama-3.1-8B-Instruct && 68.40 & 56.78 & 34.48 && 57.17 & 44.58 & 7.59  && 50.79 & 52.84 & 9.46  && 52.61 & 55.52 & 5.23  && 64.71 & 50.52 & 20.86 \\
        & Qwen2.5-7B            && 91.57 & 85.41 & 73.95 && 87.97 & 87.02 & 61.81 && 85.17 & 77.02 & 27.44 && 79.09 & 80.58 & 19.86 && 89.55 & 78.87 & 54.89 \\
        & Qwen3-8B              && \textbf{97.56} & \textbf{92.99} & \textbf{87.69} && \textbf{94.09} & 93.63 & 80.39 && \textbf{95.27} & 80.99 & 31.34 && 83.62 & 80.56 & 22.95 && \textbf{94.42} & 83.77 & 62.83 \\
        & Qwen3-32B             && 94.01 & 91.24 & 81.71 && 92.41 & \textbf{93.92} & \textbf{81.22} && 92.43 & \textbf{81.44} & \textbf{37.33} && \textbf{87.11} & \textbf{82.30} & \textbf{26.04} && 92.97 & \textbf{84.42} & \textbf{64.02} \\
        \midrule

        \multirow{4}{*}{\textbf{\begin{tabular}[l]{@{}l@{}}ASR + \\ LLM\end{tabular}}} 
        & Llama-3.1-8B-Instruct && 68.51 & 45.29 & 27.61 && 56.96 & 37.47 & 5.91  && 49.84 & 45.42 & 5.99  && 44.95 & 49.53 & 1.74  && 63.88 & 42.97 & 15.90 \\
        & Qwen2.5-Omni  && 93.02 & 79.69 & 63.86 && 82.49 & 76.74 & 47.05 && 80.44 & 73.07 & 18.61 && 40.07 & 66.76 & 7.32  && 84.02 & 73.85 & 48.80 \\
        & Qwen3-8B              && \textbf{97.67} & 78.77 & \textbf{68.07} && \textbf{92.83} & 76.58 & 49.58 && \textbf{94.64} & 73.95 & 17.98 && 86.41 & 74.40 & 13.94 && \textbf{94.29} & 72.97 & 47.27 \\
        & Qwen3-32B             && 94.90 & \textbf{78.90} & 64.63 && 92.19 & \textbf{77.94} & \textbf{51.27} && 91.17 & \textbf{78.49} & \textbf{23.77} && \textbf{86.76} & \textbf{78.70} & \textbf{16.63} && 92.66 & \textbf{76.01} & \textbf{48.96} \\
        \midrule
        
        \multirow{5}{*}{\textbf{LALM}} 
        & Qwen2.5-Omni          && 93.46 & 81.67 & 65.19 && 80.38 & 76.45 & 47.68 && 73.50 & 69.56 & 17.35 && 46.69 & 68.35 & 8.01  && 83.59 & 73.84 & 49.03 \\
        & Qwen3-Omni            && 86.47 & 81.69 & 72.06 && 86.08 & 82.70 & 60.55 && 49.21 & 71.30 & 8.20  && 46.34 & 73.76 & 13.24 && 78.83 & 77.23 & 55.07 \\
        & Gemini-2.5-Flash      && 98.56 & 85.61 & 77.17 && 90.72 & 80.81 & 55.91 && 81.07 & 82.04 & 35.02 && 67.94 & 81.80 & 25.09 && 91.45 & 82.60 & 64.40 \\
        & Gemini-2.5-Pro        && 95.23 & 87.12 & 78.05 && 90.08 & 83.24 & 63.29 && 88.64 & \textbf{85.06} & \textbf{47.95} && 73.52 & 81.62 & \textbf{35.19} && 91.25 & 84.21 & \textbf{69.00} \\
        & GPT-4o-Audio          && \textbf{98.89} & \textbf{88.12} & \textbf{80.27} && \textbf{92.83} & \textbf{84.84} & \textbf{64.56} && \textbf{89.27} & 84.12 & 35.96 && \textbf{77.00} & \textbf{83.53} & 30.31 && \textbf{94.08} & \textbf{85.40} & 68.30 \\
        \bottomrule
    \end{tabular}
    }
\end{table*}

\begin{table*}[!t]
    \centering
    \renewcommand{\arraystretch}{1.1}
    \footnotesize
    \caption{Error case analysis of GPT-4o-Audio on the Level 3 test set. \textcolor{RoyalBlue}{Blue} text highlights hallucinated dates or improperly formatted natural language responses produced by the model.}
    \resizebox{\textwidth}{!}{
    \begin{tabular}{l p{4cm} p{5cm} p{5cm}}
        \toprule
        \textbf{ID} & \textbf{User Speech Request} & \textbf{Model Prediction} & \textbf{Ground Truth} \\
        \midrule
        
        Case 1 & Hey, could you set a reminder for me to pick up the dry cleaning this Friday? & 
        [\texttt{add\_reminder}( content = "pick up the dry cleaning", time = "\textcolor{RoyalBlue}{2025-12-12 00:00:00}" )] & 
        [\texttt{add\_reminder}(content = "pick up the dry cleaning", time = "\textcolor{RoyalBlue}{NAN}" )] \\
        \midrule

        Case 2 & Um, could you maybe change the destination for my train order? I'd like the new destination to be Geelong. & 
        \textcolor{RoyalBlue}{Please provide the order ID for the train order you'd like to update with the new destination.} & 
        \textcolor{RoyalBlue}{[\texttt{update\_train\_order\_dest} (order\_id = "NAN", destination = "Geelong")]} \\
        \bottomrule
    \end{tabular}}
    \label{tab:gpt4o_error_cases}
    \vspace{-5pt}
\end{table*}

\section{Experiments}

Experiments are organized into three components: \textbf{1) SLU-SFC Comparative Analysis}: compare the performance of models when processing identical semantic datasets under traditional SLU and our proposed SFC perspective. \textbf{2) Benchmark Study}: evaluate LLMs and LALMs' performance on SFC-Bench across varying levels of difficulty. \textbf{3) Post-Training}: enhance the SFC capabilities of LALMs by integrating SFT and RL techniques, while maintaining models' general capabilities.

\subsection{SLU vs SFC}


We initially conducted comparative experiments using the \textbf{Level 1 subset} of SFC-Bench. These experiments involved both ICL and SFT across LLMs Qwen2.5-7B and Qwen3-8B \cite{yang2025qwen3}, and LALM Qwen2.5-Omni-7B \cite{xu2025qwen3}. For LLMs, the inputs consisted of textual user queries, whereas the LALM received the corresponding speech input with the same context. All models were deployed by vLLM \cite{kwon2023efficient} framework to accelerate inference. In the SFT phase, we utilized the Llama-Factory \cite{zheng2024llamafactory} training framework and applied Low-Rank Adaptation (LoRA) \cite{hu2022lora} for parameter-efficient fine-tuning.

\subsubsection{Quantitative Analysis} \textbf{SFC consistently outperforms the traditional SLU, across both text and audio modalities}. As shown in Table \ref{sfc-tab:combined_results}, SFC demonstrates an improvement of 5–18\% in both audio and text modalities. These gains are primarily attributed to the more granular and rigorous function definitions inherent in SFC, which enhance the accuracy of \textit{parameter value} (analogous to the \textit{slot filling} in traditional SLU), thereby boosting the overall performance. In contrast, SFC exhibits no significant advantage in \textit{function name} (equivalent to \textit{intent classification} in SLU). The reason is that both function name and intent classification are relatively straightforward classification problems that do not necessitate complex rule constraints.

\subsubsection{Qualitative Analysis} \textbf{Core advantage of SFC over traditional SLU lies in its more structured rule definitions.} SFC utilizes explicit schemas for tools and parameters; for any given tool, the parameter rules are fixed. For instance, as shown in Table \ref{tab:qualitative_comparison}, the \textit{delete\_reminder} tool is strictly defined with two parameters: \textit{content} and \textit{time}. In traditional SLU, however, the rules for intent classification and slot filling are often decoupled. For a specific intent, the associated slots are not explicitly constrained, and the extraction results depend entirely on the semantic information present in the user’s query. This higher degree of freedom frequently leads to elevated error rates.


\textbf{Structured SFC definitions provide the model with higher generation confidence.} We further compared the output confidence of the Qwen2.5-Omni-7B model during inference for both SFC and SLU tasks. As illustrated in the Table \ref{fig:log_prob}, the model exhibits significantly higher \textit{Log Probabilities} when performing the SFC task, indicating superior decoding confidence, which ultimately leads to a higher task success rate.

\subsection{Benchmark}


Consistent with the comparative experiments, all open-source models were deployed using the vLLM \cite{kwon2023efficient} framework to accelerate inference. For the pipeline systems comprising an ASR model integrated with LLMs, the widely adopted Whisper-Large-V3-Turbo \cite{radford2023robust} was utilized as the unified ASR component, achieving a Word Error Rate (WER) of 13.69\%. 


\begin{table*}[tbp]
    \centering
    \renewcommand{\arraystretch}{1.1}
    \setlength{\tabcolsep}{3pt}
    \caption{Post-training results and ablation studies. \textbf{SpokenFC-7B} denotes our final model trained with the GRPO algorithm, fine-grained rewards, FP16 precision, and large rollouts ($N=32$). Best results are in \textbf{bold}.}
    \label{sfc-tab:comprehensive_results}
    \resizebox{\textwidth}{!}{%
    \begin{tabular}{l ccccc ccccc ccc}
        \toprule
        \multirow{2}{*}{\textbf{Model / Variant}} & 
        \multicolumn{5}{c}{\textbf{SFC Test-ID}} & 
        \multicolumn{5}{c}{\textbf{SFC Test-OOD}} & 
        \textbf{CV 15} & \textbf{MMSU} & \textbf{API-Bank} \\
        
        \cmidrule(lr){2-6} \cmidrule(lr){7-11} \cmidrule(lr){12-12} \cmidrule(lr){13-13} \cmidrule(lr){14-14}
        
        & Level 1 & Level 2 & Level 3-1 & Level 3-2 & Overall & 
          Level 1 & Level 2 & Level 3-1 & Level 3-2 & Overall & 
          WER ($\downarrow$) & Acc ($\uparrow$) & Overall ($\uparrow$) \\
        \midrule
        
        Qwen2.5-Omni-7B & 63.86 & 47.05 & 18.61 & 7.32 & 48.80 & 68.54 & 51.06 & 24.60 & 3.85 & 56.16 & 8.44 & 62.54 & 54.84 \\
        GPT-4o-Audio & 80.27 & 64.56 & 35.97 & 30.31 & 68.30 & 71.65 & 64.89 & 49.21 & 26.92 & 68.09 & -- & -- & -- \\
        Gemini-2.5-Pro & 78.05 & 63.29 & 47.95 & 35.19 & 69.00 & 73.52 & 67.02 & 57.14 & 38.46 & 71.60 & -- & -- & -- \\
        \textbf{SpokenFC-7B (Proposed)} & \textbf{81.60} & \textbf{71.10} & 49.21 & 40.42 & 74.15 & \textbf{76.95} & \textbf{69.15} & 65.05 & 38.46 & \textbf{75.10} & 8.30 & \textbf{62.58} & \textbf{64.52} \\
        \midrule

        \multicolumn{14}{l}{\textit{Ablation Variants}} \\
        \quad SFT & 80.38 & 66.67 & 47.95 & 36.93 & 71.90 & 74.45 & 62.77 & 65.08 & \textbf{46.15} & 73.93 & 8.49 & 62.50 & 55.58 \\
        \quad RL w/ EM Reward & 81.42 & 70.23 & 47.25 & 36.67 & 73.08 & 76.34 & 68.21 & 64.42 & 36.97 & 73.24 & 8.35 & 62.30 & 63.85 \\
        \quad RL w/ BF16 & 81.37 & 69.41 & 44.79 & 37.28 & 72.83 & \textbf{76.95} & 67.02 & 63.49 & 36.54 & 74.32 & 8.32 & 62.44 & 63.97 \\
        \quad RL w/ $N=8$ & 77.12 & 60.33 & 40.50 & 27.15 & 67.40 & 67.10 & 60.45 & 52.20 & 24.15 & 66.80 & 8.35 & 62.20 & 60.50 \\
        \quad RL w/ $N=16$ & 80.31 & 66.27 & 44.09 & 33.21 & 71.24 & 73.53 & 65.42 & 59.78 & 30.67 & 71.44 & 8.30 & 62.38 & 62.14 \\
        \quad RL w/ $N=64$ & 81.37 & 70.25 & \textbf{50.47} & \textbf{42.51} & \textbf{74.38} & 76.32 & 68.09 & \textbf{66.46} & 40.18 & 75.03 & \textbf{8.29} & 62.50 & 64.43 \\
        
        \bottomrule
    \end{tabular}%
    }
\end{table*}

\subsubsection{Quantitative Analysis} \textbf{Speech-based function calling is significantly more challenging than text-based tasks, primarily due to ASR error propagation.} As shown in Table \ref{tab:test_id_en}, comparing Oracle text with pipelines reveals a substantial performance drop. For instance, Qwen3-32B falls from 64.02\% accuracy to 48.96\% when transitioning from text to speech. ASR misrecognitions of proper nouns and numerical values serve as the primary bottleneck, severely undermining downstream intent understanding.

\textbf{Task complexity serves as the primary determinant of model performance.} All evaluated architectures exhibit significant performance degradation as the difficulty level increases. Specifically, even GPT-4o-Audio and Gemini-2.5-Pro experience a sharp decline in overall accuracy, dropping from 80.27\% at Level 1 to 30.31\% at Level 3-2, which underscores the limitations of current models in handling \texttt{NAN} parameter identification and multi-turn semantic completion.

\textbf{Open-source LALMs show impressive progress but still lag behind closed-source models in high-level scenarios.} While Qwen3-Omni is competitive with Gemini-2.5-Pro and GPT-4o-Audio in Level 1 and 2 tasks, a significant performance gap emerges in Level 3. Closed-source models demonstrate superior robustness and generalization when handling the complex instructions and unseen domains characteristic of high-level tasks.


\subsubsection{Qualitative Analysis} To further investigate the failure modes of LALMs in SFC tasks, Table~\ref{tab:gpt4o_error_cases} presents typical error cases on the high-difficulty test set. Even for GPT-4o-Audio, processing spoken interactions still presents the following critical challenges: \textbf{1) Model Hallucination.} As demonstrated in case 1, when user queries contain vague temporal expressions (e.g., ``\textit{this Friday}''), the model tends to hallucinate a specific time (e.g., ``\textit{2025-12-12 00:00:00}'') rather than filling \textit{NAN} as required when specific parameters are missing. This over-compensation behavior can lead downstream systems to execute incorrect instructions. \textbf{2) Instruction Following Failure.} As illustrated in case 2, when faced with scenarios requiring the completion of critical missing parameters (e.g., Order ID), the model occasionally reverts to a traditional chatbot mode. It provides a natural language inquiry instead of outputting a structured function calling sequence as required.

\subsection{Post-Training and Ablation Study}

\begin{figure}[t!]
    \centering
    \begin{minipage}[b]{0.205\textwidth}
        \centering
        \includegraphics[width=\textwidth]{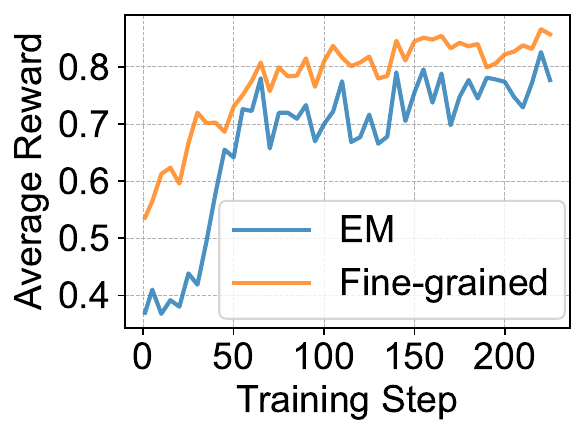} 
        \vskip -0.1in
        \caption{Ablation study on reward types.}
        \label{fig:ablation-reward}
    \end{minipage}%
    \hfill
    \begin{minipage}[b]{0.26\textwidth}
        \centering
        \includegraphics[width=\textwidth]{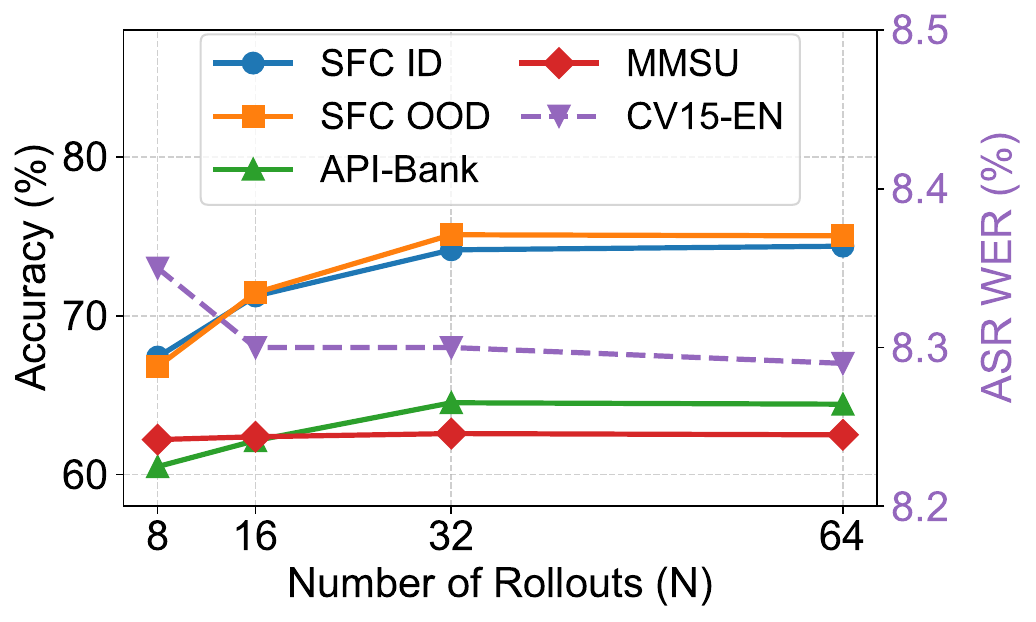} 
        \vskip -0.1in
        \caption{Ablation study among different rollouts.}
        \label{fig:ablation-sample-size}
    \end{minipage}
    \vspace{-10pt}
\end{figure}
We employ Llama-Factory \cite{zheng2024llamafactory} for SFT and MS-SWIFT \cite{zhao2025swift} for RL. All experiments are based on Qwen2.5-Omni-7B, fine-tuning only the LLM backbone via LoRA while freezing the vision encoder, audio encoder, and talker modules. Models are trained for one epoch on Nvidia 8$\times$A800 GPUs. To evaluate generalization, we test on ASR with CommonVoice15-en \cite{ardila2020common}, audio reasoning on MMSU \cite{wang2025mmsu}, and text function calling on API-Bank \cite{li2023api}, level 1 subset. Experimental results are in Table \ref{sfc-tab:comprehensive_results}. 



\textbf{SpokenFC-7B model, fine-tuned via synthetic training data, demonstrates substantial performance improvements in both Test-ID and Test-OOD SFC tasks, as well as text-based function calling.} Notably, it surpasses GPT-4o-Audio in SFC performance, and also achieves marginal gains in ASR and audio reasoning tasks compared to the base model, further validating the cross-modal generalization capability of the post-training.


\textbf{RL exhibits enhanced generalization and transferability over SFT.} Although SFT facilitates notable improvements in SFC tasks through memorization, it fails to extend effectively to analogous text-based function call tasks. In contrast, RL encourages the model to internalize the fundamental logic of semantic parsing and tool utilization, leading to performance enhancements in both speech and text domains. Moreover, whereas SFT leads to a slight regression in ASR and Audio Reasoning capabilities, RL training successfully addresses the issue of catastrophic forgetting, actually bolstering performance in these foundational tasks.

\textbf{Fine-grained rewards address the sparse reward challenge in early-stage RL by providing dense signals, as shown in Figure \ref{fig:ablation-reward}.} This granular feedback is particularly advantageous for complex tasks with long reasoning chains (e.g., Level 3-2); by reducing variance in policy optimization, it ensures a smoother and more robust training process while improving the model's final performance.

\textbf{FP16 outperforms BF16 by leveraging its higher mantissa precision to mitigate the training-inference mismatch \cite{qi2025defeating}.} This increased precision reduces cumulative rounding errors during long JSON sequence generation and stabilizes training gradients to prevent model collapse. Consequently, FP16 achieves a superior balance between task-specific optimization and the maintenance of general capabilities.

\textbf{Increasing the number of rollouts ($N$) significantly boosts exploration efficiency and provides more accurate baseline estimates for advantage functions, yet it follows the law of diminishing returns.} While a larger model can further maximize in-domain performance, excessive sampling risks overfitting to specific training patterns, thereby compromising OOD generalization.

\section{Conclusion}

This work introduces Spoken Function Calling as a novel perspective for task-oriented spoken semantics extraction, to evolve beyond traditional closed-set SLU. Leveraging structured rule definitions, SFC overcomes the limitations of traditional SLU in handling open-domain tasks with ICL. Through the dataset construction and empirical evaluations, this study demonstrates that SFC significantly improves semantic understanding accuracy compared to traditional SLU for LLMs and LALMs, and SFC performance can be further enhanced through RL post-training with fine-grained rewards. Future work will be directed toward dataset scaling and post-training techniques to further refine the model's capabilities.




\noindent\textbf{Acknowledgments.}
This work was supported by National Natural Science Foundation of China  (No. U23B2018), the Science and Technology Innovation (STI) 2030-Major Project (2022ZD0208700), Yangtze River Delta Science and Technology Innovation Community Joint Research Project (2024CSJGG1100), and Alibaba Research Intern Program.

\bibliographystyle{ACM-Reference-Format}
\bibliography{main}


\end{document}